\documentclass[11pt]{article}

\usepackage[final]{acl}

\usepackage{times}
\usepackage{latexsym}

\usepackage[T1]{fontenc}

\usepackage[utf8]{inputenc}

\usepackage{microtype}

\usepackage{inconsolata}

\usepackage{graphicx}

\usepackage{booktabs}
\usepackage{enumitem}

\usepackage{listings}

\usepackage[most]{tcolorbox}
\usepackage{xcolor}

\title{Detecting AI-Generated Content on Social Media\\ 
with Multi-modal Language Models}

\author{
 \textbf{Chenyang Yang\textsuperscript{1}\thanks{Work done at Meta.}},
 \textbf{Shen Yan\textsuperscript{2}},
 \textbf{Yibo Yang\textsuperscript{2}},
 \textbf{Litao Hu\textsuperscript{2}},
 \textbf{Yuchen Liu\textsuperscript{2}},
 \textbf{Yuan Zeng\textsuperscript{2}},\\
 \textbf{Hanchao Yu\textsuperscript{2}},
 \textbf{Yinan Zhu\textsuperscript{2}},
 \textbf{Sumedha Singla\textsuperscript{2}},
 \textbf{Brian Vanover\textsuperscript{2}},
 \textbf{Huijun Qian\textsuperscript{2}},
\\
 \textbf{Zihao Wang\textsuperscript{2}},
 \textbf{Fujun Liu\textsuperscript{2}},
 \textbf{Aashu Singh\textsuperscript{2}},
 \textbf{Jianyu Wang\textsuperscript{2}},
 \textbf{Xuewen Zhang\textsuperscript{2}}
\\
\\
 \textsuperscript{1}Carnegie Mellon University,
 \textsuperscript{2}Meta
}

\begin{document}
\maketitle
\begin{abstract}
Generative AI has enabled the creation of photorealistic images and videos that are increasingly disseminated on social media, often used for spam, misinformation, manipulation, and fraud. 
Existing AI-generated content (AIGC) detection methods face challenges including poor generalization to new generation models, reliance on single modalities, and lack of interpretable explanations. 
We present our pipeline that mitigates these issues by continuously curating diverse multi-modal social media data and training a compact vision-language model for detection and explanation. 
Our model achieves state-of-the-art detection performance on public benchmarks and demonstrates robust detection and explanation capabilities on internal social media datasets across multiple platforms. 
We deployed our model for post recommendation on social media platforms and observed positive downstream impacts on user engagement, demonstrating that it is feasible to perform effective AIGC detection in dynamic, real-world social media environments.
\end{abstract}

\newcommand\circleone{\ding{192}}
\newcommand\circletwo{\ding{193}}
\newcommand\circlethree{\ding{194}}
\newcommand\circlefour{\ding{195}}
\newcommand\circlefive{\ding{196}}
\newcommand\circlesix{\ding{197}}
\newcommand\circleseven{\ding{198}}
\newcommand\circleeight{\ding{199}}

\newcommand{\ourmodel}[0]{\textsc{IFM-AIGCSpotter-3b}}

\definecolor{mybgcolor}{HTML}{E6F2F0}    %
\definecolor{myframecolor}{HTML}{FF8552} %
\definecolor{mytitlecolor}{HTML}{2C7873} %

\newcommand{\nbc}[3]{
 {\colorbox{#3}{\bfseries\sffamily\scriptsize\textcolor{white}{#1}}}
 {\textcolor{#3}{\sf$\blacktriangleright$\textit{#2}$\blacktriangleleft$}}
 }

 \newcommand\todo[1]{\nbc{TODO}{#1}{red}}
\newcommand{\cyang}[1]{\nbc{CY}{#1}{teal}}

\lstset{
  language=Python,
  keywords={},
  commentstyle=\textbf,
  stringstyle=\text,
  extendedchars=false,
  basicstyle=\ttfamily,
  columns=fullflexible,
  frame=single,
  breaklines=true,
  breakatwhitespace=true,
  breakindent=2\dimen0,
  literate={'}{{\textquotesingle}}1 {`}{{\textasciigrave}}1 {"}{{\textquotedbl}}1,
}

\newtcolorbox{nicebox}{
  colback=gray!5,
  colframe=gray!60,
  boxrule=0.6pt,
  arc=4pt,
  left=8pt,
  right=8pt,
  top=6pt,
  bottom=6pt
}

\section{Introduction}
Generative AI is increasingly capable of generating photorealistic images and videos disseminated over social media~\cite{rombach2022high, brooks2024video}.
AI-generated contents are being used to generate spammy clickbait~\cite {forbes2024clickbait}, to
create and spread misinformation~\cite{associatedpress2024trump}, to manipulate and influence behaviors~\cite{tarsney2025deception}, as well as to harass, scam, and fraud users~\cite{wired2024aipoweredscams}.

This has motivated a large body of work on AI-generated content (AIGC) detection:
Dedicated models have been developed to detect deepfakes~\cite{frank2020leveraging}, AI-generated arts~\cite{rahman2023artifact}, and lately photorealistic images~\cite{yan2024sanity}.
However, the existing approaches generally suffered from three problems:
(1) They are trained on static datasets and \textbf{do not generalize well} to images generated by newer models~\cite{epstein2023online},
(2) they only rely on information from one single modality for detection~\cite[e.g.,][]{frank2020leveraging, yan2024sanity}, and can not utilize rich \textbf{multi-modal signals} from social media posts, and
(3) they mostly do not provide human-readable explanations, that make their judgments \textbf{hard to interpret} for humans.

In this work, we share our approach to detect and explain social media posts with AI-generated content at Meta, aiming to mitigate the above three problems (Section~\ref{sec:method}).
First, we develop a dedicated data curation pipeline, that allows us to continuously curate new high-quality AIGC data from various social media platforms.
Second, we adopt the latest developments in multi-modal foundation models~\cite{grattafiori2024llama}, and train a small yet effective vision language model (\ourmodel) that detects and explains social media posts with AI-generated content.

We evaluated our approach through two sets of experiments: 
We first benchmark our approach against public datasets (Section~\ref{sec:experiment-benchmark}), showing our model achieve state-of-the-art performance on par with other open-sourced models.
We next evaluate our trained model on an internal dataset (Section~\ref{sec:experiment-internal}), demonstrating our model's ability to accurately detect AI-generated contents and provide useful explanations.
We conducted further ablations and deployed our model for post recommendation at social media platforms.
Overall, we demonstrate that it is possible to detect AI-generated content effectively from a large quantity of, diverse, and continuously evolving social media posts.

\section{Related Work}
A substantial body of work studies AI-generated content (AIGC) detection. Early approaches exploit low-level spatial or frequency artifacts~\cite{mccloskey2018detecting, frank2020leveraging}, while lightweight CNNs trained on GAN datasets capture generator-specific cues~\cite{wang2020cnn, rossler2019faceforensics++}. The rise of diffusion models has spurred new detectors, including CLIP-based methods that show improved cross-generator generalization~\cite{cozzolino2024raising}. To cope with rapidly evolving generators, recent work frames detection as a continual-learning problem, studying online or continuous adaptation~\cite{epstein2023online, tassone2024continuous}.

\begin{figure}[t]
    \centering
    \includegraphics[trim=130pt 260pt 1050pt 150pt, clip, width=\linewidth]{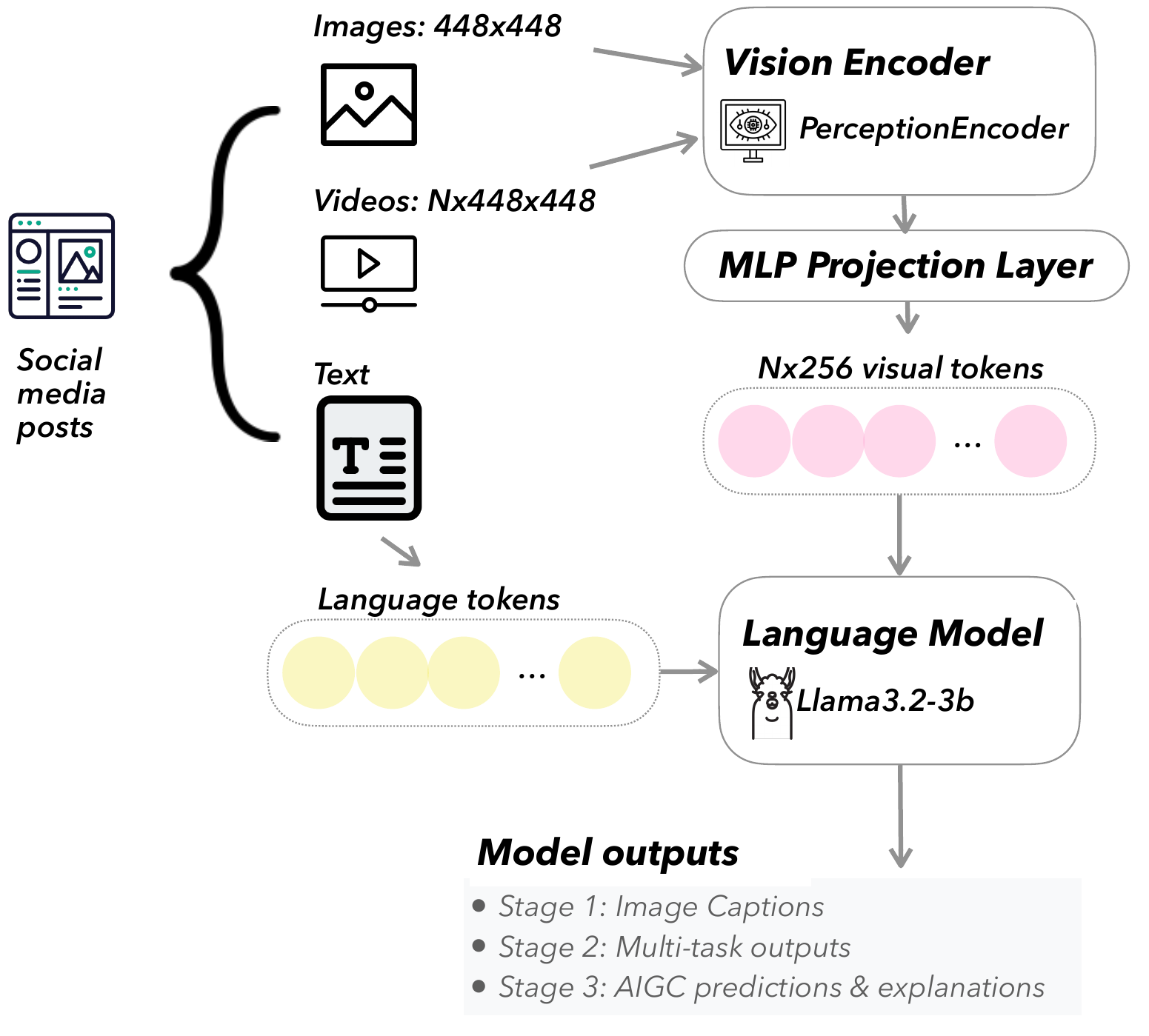}
    \caption{Overview of our LLaVA-based model architecture. Our model processes  images and videos with PerceptionEncoder into visual tokens, which are fed into a language model together with language tokens.
    }
    \label{fig:model_arch}
\end{figure}

\begin{figure*}[ht]
    \centering
    \includegraphics[trim=120pt 350pt 600pt 100pt, clip, width=\linewidth]{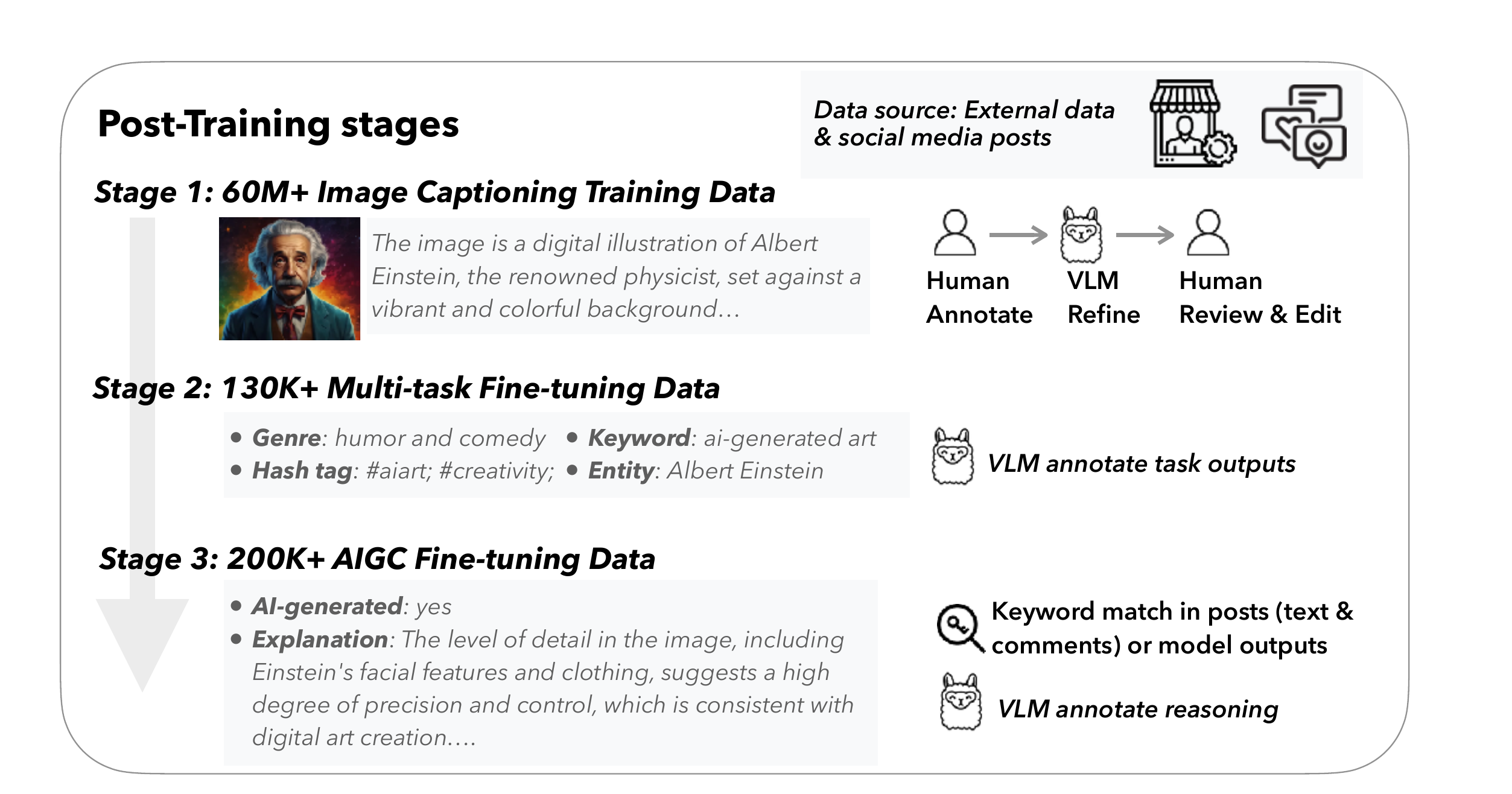}
    \caption{Overview of the data curation and post-training pipeline. The model is initialized with pretrained weights of \texttt{Llama3.2-3B-Instruct}~\cite{grattafiori2024llama} and Perception Encoder~\cite{bolya2025perception}. 
    It is further post-trained on three curated data categories: 
    (1) Image Captioning data from Shutterstock and social media, including human-annotated high-quality captions; 
    (2) Multi-task Fine-tuning data for enhanced multimodal understanding with tasks like named entity recognition and hashtag generation; 
    (3) AIGC Fine-tuning data for AI-generated content detection and explanation, curated via heuristic labeling and continuous automated collection.}
    \label{fig:overview}
\end{figure*}

\paragraph{Vision-language models.}
Large pretrained VLMs provide a much stronger multimodal prior for AIGC detection. 
Prior work explores linear probes or prompt tuning on frozen CLIP backbones~\cite{ojha2023towards, keita2024fidavl}, in-context prompting of multimodal LLMs~\cite{jia2024can, ye2024loki}, and fine-tuning MLLMs to reason about visual semantics~\cite{liu2024forgerygpt, wen2025spotfakelargemultimodal}.

\paragraph{In-the-wild benchmarks.}
Recent datasets better reflect real-world distribution shifts. MiRAGeNews~\cite{huang2024miragenews} pairs Midjourney images with synthetic captions, revealing weaknesses on unseen publishers. Deepfake-Eval-2024~\cite{chandra2025deepfake} and BusterX++~\cite{wen2025busterx++} collect recent social-media fakes, causing large performance drops for state-of-the-art detectors. These benchmarks highlight the need for multimodal reasoning and continual adaptation.

\paragraph{Discussion.}
Most existing detectors are specialized to particular generators or rely solely on visual artifacts, and current benchmarks largely omit social context.
Our work addresses these limitations by continuously collecting image-text pairs with social signals and training a unified VLM that leverages both visual and contextual cues for robust, real-time AIGC detection.

\section{Method}
\label{sec:method}

In this section, we provide a detailed description of our method.
We first outline our model architecture, and then present our data curation pipelines and their corresponding model training recipes.

\subsection{Model Architecture}

We follow the general architecture of LLaVA~\cite{liu2023visual}, a widely adopted architecture for vision language modeling such as in Qwen2.5-VL~\cite{bai2025qwen2} and InternVL-2.5~\cite{chen2024expanding}.
The LLaVA architecture has three major components: a Vision Encoder, a Language Model, and an MLP projection layer that connects the Vision Encoder and Language Model, as demonstrated in Figure~\ref{fig:model_arch}.

\paragraph{Implementation.}
To ensure high inference efficiency, we choose a model of small size, \texttt{Llama3.2-3B-Instruct}~\cite{grattafiori2024llama} as our Language Model.
For Vision Encoder, we choose Perception Encoder~\cite{bolya2025perception} with 300M parameters, as it demonstrates strong zero-shot classification results compared to other vision encoders~\cite{bolya2025perception}. 
To improve inference efficiency, we crop the image to a resolution of $448\times 448$, and adopt an adaptive pooling strategy to produce a total of 256 output visual tokens per image.
For the projection layer, we use a single linear layer with zero-weight initialization.

\subsection{Data Curation}
We initialize our model with the pretrained weights and train our model on a wide range of curated image-text data.
The data can be categorized into three parts as illustrated in Figure \ref{fig:overview}:

\begin{itemize}[noitemsep, topsep=0pt]
    \item \textbf{Image Captioning} data to align the image-text modality.
    \item \textbf{Multi-task Fine-tuning} data to improve multimodal content understanding capabilities.
    \item \textbf{AIGC Fine-tuning} data to adapt the model to the downstream AIGC detection and explanation tasks.
\end{itemize}

\paragraph{Image Captioning Data.}
We collected a total of 60M+ image captioning data from three sources: 

\begin{itemize}[noitemsep, topsep=0pt]
    \item \textbf{Shutterstock Split}: We leveraged a substantial subset of the Shutterstock dataset, comprising 60 million images, after rigorous cleaning. The initial round of data cleaning removed low-quality data, including non-ASCII-coded characteristics, repeating words, incomplete sentences, etc.
    We then used \texttt{Llama-3.2-90B} to generate elaborate captions for the filtered data, while providing original captions for grounding.
    \item \textbf{Social Media Split}: We sampled 10M images and 2.2M videos from public posts on social media platforms based on post engagement. 
    We then generated captions for these images with \texttt{Llama4-Scout}.
    \item \textbf{High-quality Social Media Split}: For a subset of 164K social media images, we have humans first annotate captions, including visual details, OCR, and a summary. 
    These captions are further refined by \texttt{Llama4-Scout} and then reviewed and edited by human annotators.
\end{itemize}

\paragraph{Multi-task Fine-tuning Data.}
In this stage, we collected data for various downstream tasks that improve model's multi-modal understanding capabilities.
We included tasks like named entity recognition, keyword generation, hashtag generation, and genre classification.
We sampled 130K images from the \textbf{Social Media Split} and generated task-specific labels with \texttt{Llama4-Maverik}.
Note that in this stage, the training data often contains signals related to AI-generated content (e.g., in \textit{generated} keywords or hashtags), which bootstraps the model to obtain some initial knowledge for later AIGC detection and explanation. 

\paragraph{AIGC Fine-tuning Data.}
In the final stage, we collected 200K samples for AIGC fine-tuning.
We sampled images from social media posts and labeled them using simple heuristics:
Posts are labeled as AIGC if their text or comments explicitly mention AIGC-related keywords based on a curated vocabulary (e.g., \textit{``ai-generated,'' ``Midjourney,'' ``AI art''}), or if the multi-task model from the previous stage predicts an AIGC-related keyword or hashtag; all others are labeled as non-AIGC. 
We further cleaned the data by clustering them with embeddings and keeping the top clusters with human confirmation.
This results in 30K AIGC and 170K negative samples, with explanations annotated with \texttt{Llama4-Maverik}. 
This stage runs continuously, enabling data collection and model re-training on emerging AI-generated content.

\subsection{Post-training Recipe}
Our post-training pipelines consist of three stages:

\begin{itemize}[noitemsep, topsep=0pt]
    \item \textbf{Stage 1 (Image Captioning)}: This initial stage equips our model with foundational image understanding capabilities.
    \item \textbf{Stage 2 (Multi-task Fine-tuning):} This stage improves our model's multimodal content understanding capabilities and helps prepare data for the final stage.
    \item \textbf{Stage 3 (AIGC Detection Fine-tuning)}: This final stage exclusively focuses on the ability to detect and explain AIGC.
\end{itemize}

\paragraph{Stage 1 (Image Captioning).}
In the first stage, our goal is to equip our model with foundational image understanding capabilities by aligning the text-image modality.
We first freeze both Language Model and Vision Encoder, and exclusively train the MLP projection layer with Shutterstock Split of image captioning data.
We use an effective batch size of 2048 and a learning rate of 0.002 using a Cosine Scheduler with 200 warm-up steps.

We then unfreeze the Vision Encoder and train on the \textbf{Social Media Split}. 
Finally, we train the model end-to-end with the small \textbf{High-quality Social Media Split}.
We use an effective batch size of 128 and 64 respectively for the last two parts of Stage 1 training, and a learning rate of 4e-5.

\paragraph{Stage 2 (Multi-task Fine-tuning).}
The goal of the second stage is to improve model's multi-modal content understanding capabilities. 
We fine-tuned the model from stage 1 on the curated \textbf{Multi-task Fine-tuning Data}.
We use an effective batch size of 256, a learning rate of 4e-5, and Adam 8-bit for optimization.

\paragraph{Stage 3 (AIGC Detection Fine-tuning).}
The goal of the final stage is to improve our model's capability to detect and explain posts with AI-generated content.
We performed standard supervised fine-tuning on the curated \textbf{AIGC Fine-tuning Data}.
During the fine-tuning, we randomly drop 90\% of the text information, such that the trained model does not solely rely on the textual clues but also learns from related features in the images.

\paragraph{Discussion.}
Our training pipeline produces a designated model, \ourmodel, that understands multi-modal social media posts well.
Different from existing vision language models~\cite{grattafiori2024llama, bai2025qwen2}, \ourmodel{} is exposed to a wide range of images posted on social media, which provides strong signals of the latest trend of AIGC, laying a foundation for detection and explanation.

\begin{table*}[ht]
    \centering
    \footnotesize
    \begin{tabular}{lcccc}
        \toprule
        \textbf{Method} & \multicolumn{2}{c}{\textbf{FakeClue (in-distribution)}} & \multicolumn{2}{c}{\textbf{LOKI (out-of-distribution)}} \\
        \cmidrule(lr){2-3} \cmidrule(lr){4-5}
        & \textbf{Acc $\uparrow$} & \textbf{F1 $\uparrow$} & \textbf{Acc $\uparrow$} & \textbf{F1 $\uparrow$} \\
        \midrule
        \textbf{Prompting only} \\
        Deepseek-VL2-small & 0.404 & 0.542 & 0.253 & 0.387 \\
        Deepseek-VL2 & 0.475 & 0.541 & 0.431 & 0.392  \\
        InternVL2-8B & 0.506 & 0.490 & 0.526 & 0.340 \\
        InternVL2-40B & 0.507 & 0.463 & 0.507 & 0.376 \\
        Qwen2-VL-7B & 0.457 & 0.592 & 0.571 & 0.350 \\
        Qwen2-VL-72B & 0.578 & 0.565 & 0.554 & 0.409 \\
        GPT-4o & 0.474 & 0.420 & 0.634 & 0.572 \\
        \midrule
        \textbf{Detection-only training} \\
        Qwen2.5-VL-3b & 0.950 & 0.962  & 0.767 & 0.809  \\
        Gemma3-4b-it & 0.946 & 0.958  & 0.748 & 0.810  \\
        Llama3.2-11b & 0.956 & 0.966  & 0.802 & 0.848  \\
        \ourmodel{} & \textbf{0.986} & \textbf{0.989}  & \textbf{0.839} & \textbf{0.870}  \\
        \midrule
        \textbf{Detection + Explanation training} \\
        Qwen2.5-VL-3b & 0.932 & 0.948 & 0.739 & 0.792   \\
        Gemma3-4b & 0.937 & 0.951 & 0.784 & 0.824   \\
        Llama3.2-11b & 0.951 & 0.962 & 0.789 & \textbf{0.846}   \\
        FakeVLM (11b) & \textbf{0.986} & \textbf{0.981}  & \textbf{0.843} & {0.837}  \\
        \ourmodel{} & 0.971 & \textbf{0.977} & 0.816 & \textbf{0.844}   \\
        \bottomrule
    \end{tabular}
    \caption{Performance comparison of various models on FakeClue and LOKI datasets. 
    \ourmodel{} achieves top detection accuracy (0.986 on FakeClue and 0.839 on LOKI) among fine-tuned models, demonstrating strong generalization especially in out-of-distribution settings. 
    We observe that incorporating explanation generation during training slightly reduces detection accuracy, suggesting a trade-off between explainability and performance.}
    \label{tab:performance}
\end{table*}

\section{Experiments}
\label{sec:experiment}

\subsection{Detecting and Explaining AI-generated Images in Benchmarks}
\label{sec:experiment-benchmark}
We first evaluate our approach by benchmarking on public datasets and comparing it against a wide range of closed-sourced and open-sourced models.

\subsubsection{Setups}
To ensure a fair comparison, we use the model checkpoint from Stage 2 before dedicated AIGC fine-tuning. 
We then follow the setup of prior work~\cite{wen2025spotfakelargemultimodal}, fine-tuning models on the same training dataset, FakeClue~\cite{wen2025spotfakelargemultimodal}, and evaluating them on FakeClue's test split (in-distribution), as well as an independent test set LOKI~\cite{ye2024loki} (out-of-distribution).
The goal here is to understand how well different models can learn to detect AIGC, and how well they generalize to different AIGC datasets.

\paragraph{Baselines.}
We compare against three baselines: (1) seven VLMs with zero-shot prompting~\cite{wu2024deepseek, chen2024internvl, wang2024qwen2} with reported results in prior work~\cite{wen2025spotfakelargemultimodal}, (2) three open-source VLMs (Qwen2.5-VL-3b~\cite{bai2025qwen2}, Gemma3-4b, \texttt{Llama3.2-11b}~\cite{grattafiori2024llama}) fine-tuned with and without explanations, and (3) prior SOTA with explanation, FakeVLM~\cite{wen2025spotfakelargemultimodal}.

\paragraph{Datasets.}
We evaluate all approaches on two datasets:
The test split of FakeClue~\cite{wen2025spotfakelargemultimodal} and an independent test set LOKI~\cite{ye2024loki}, to test the approaches' abilities to detect AIGC both in in-distribution and out-of-distribution settings.
The datasets cover a wide range of AIGC, from deepfakes, synthesized images of animals, humans, objects, and scenes, to AI-generated documents and satellite images.

We report accuracy and F1 scores on both datasets.
Additional ablation results can be found at Appendix~\ref{appendix:ablation}.

\subsubsection{Results}
Overall, we found \ourmodel{} among the best-performing models to detect AI-generated content (Table~\ref{tab:performance}), with an accuracy of 0.986 on FakeClue and an accuracy of 0.839 on LOKI (out-of-distribution).

Breaking down the results, we first found that \textbf{existing VLMs struggle to detect AIGC correctly without specialized fine-tuning}, with the best performing model achieving < 0.6 accuracy on the FakeClue dataset.

Second, comparing different fine-tuned models, we found \textbf{fine-tuned models with a small size are able to detect AIGC reliably}.
Specifically, we found \ourmodel{} demonstrates better generalizations to LOKI under both training schemes -- it is particularly effective for images with persons (+5.4\%), animals (+6.2\%), and scenes (+3.5\%), but less so for documents (-0.3\%). 
We believe this largely reflects our model’s training data, where documents appear much less frequently compared to persons, animals, or scenery.

Third, we found adding explanation to the training process slightly compromises detection quality, which contradicts with findings in prior work~\cite{wen2025spotfakelargemultimodal}.
We hypothesize that, adding explanation generation during training introduces a trade-off in model capacity and optimization, which lead the model to allocate resources to generating explanations at the expense of optimizing detection accuracy. 
However, we consider this is a trade-off that is potentially worth making, to introduce better explainability and user trust.

\begin{table*}[t]
    \centering
    \footnotesize
    \resizebox{\linewidth}{!}{
        \begin{tabular}{lccccccccccc}
            \toprule
            \textbf{Dataset} & \textbf{CNNSpot} & \textbf{FreDect} & \textbf{Fusing} & \textbf{GramNet} & \textbf{LNP} & \textbf{UnivFD} & \textbf{DIRE} & \textbf{PatchCraft} & \textbf{NPR} & \textbf{AIDE} & \textbf{Ours} \\
            \midrule
            Chameleon & 60.11 & 56.86 & 57.07 & 60.95 & 55.63 & 55.62 & 59.71 & 56.32 & 58.13 & 62.60 & \textbf{86.00} \\
            \bottomrule
        \end{tabular}
    }
    \caption{Performance comparison across different detection methods on the Chameleon dataset. Our method significantly outperforms existing baselines without adaptation.}
    \label{tab:results-chameleon}
\end{table*}

\begin{table}[t]
    \centering
    \footnotesize
    \begin{tabular}{lcccc}
        \toprule
        \textbf{Platform} & \textbf{Precision} & \textbf{Recall} \\
        \midrule
        Facebook & 0.888 & 0.818 \\
        Instagram & 0.883 & 0.835 \\
        Threads & 0.926 & 0.919 \\
        Overall  & 0.886 & 0.853 \\
        \bottomrule
    \end{tabular}
    \caption{Performance of our model in detecting AI-generated content (AIGC) across three platforms.
    These results demonstrate the model’s reliable detection capability across diverse social media environments.}
    \label{tab:aigc-results-platforms}
\end{table}

\subsection{Detecting AI-generated Content on Social Media}
\label{sec:experiment-internal}
Next, we evaluate \ourmodel's ability to detect AIGC on real-world social media posts. 
We use the model checkpoint from stage 3 for this evaluation.
Additional experiments on exploring the model's reasoning capabilities are reported in Appendix~\ref{sec:experiment-internal-explain}.
We also share more detailed error analysis in Appendix~\ref{appendix:error-analysis}.

\subsubsection{Internal Evaluation}
We curate test data using the same pipeline as in Section~\ref{sec:method} and report results by platform. 
We compare against two baselines: (1) an embedding-based MLP trained on visual embeddings, and (2) a lightweight adapter variant with the backbone frozen.
In addition, we run ablations to isolate key design choices: (1) removing random text drop, (2) swapping the Perception Encoder for a CLIP-based vision backbone, and (3) using 4- and 8-frame videos instead of the default 1-frame input.

\paragraph{Results.}
Overall, we found our model can reliably detect AIGC across all three platforms we look at (Table~\ref{tab:aigc-results-platforms}), and significantly improve over baselines (Table~\ref{tab:aigc-results-compare}).

For ablations, we found that random text drop is crucial for recall, as models will over-rely on the textual clues otherwise.
We also found the Perception Encoder offers a slight precision advantage over CLIP vision encoder, 
and observed significant improvement by extending the inputs to multi-frame (Table~\ref{tab:aigc-results-compare}).

\subsubsection{Generalizations to Benchmarks}
We further evaluate our trained model on public benchmarks \textit{without} further adaptation.
We choose Chameleon~\cite{yan2024sanity} as it contains highly realistic AIGC images commonly observed on social media.
We compare our method to the reported results of 10 existing methods~\cite{yan2024sanity}.

\paragraph{Results.}
We found our model achieves 86.0\% accuracy (Table~\ref{tab:results-chameleon}), outperforming all baselines by a large margin (>20\%).  
This result demonstrates that our data curation pipeline is high-quality and is able to train detectors that reliably detect realistic AIGC images without dedicated adaptation.

\subsection{Deployment in Production}

We deployed the model with 4 Nvidia H100 nodes. 
It will trigger all of the posts with views > 500 (millions per day) for global users.
We aggregated the results per content creator and identified the ones with excessively AIGC posts.
We filtered out the creators with high engagement and demoted the remaining ones for new users and marginal users.
We conducted standard online A/B testing, with the model deployed for the treatment group.

\paragraph{Results.}
After deploying the model for 6 weeks, in the 7-day backtest we noticed there is a statistically significant views gain for marginal users (+0.21\%). 
For global, the launch achieved +0.22\% Young-Adult session gains. 
This demonstrates that, our deployment strategy successfully improves users' experience on social media, resulting in higher user engagements.

\section{Discussion}

Our results demonstrate that our approach achieves state-of-the-art performance in detecting AI-generated content. 
We next discuss practical implications of our work and future challenges.

\paragraph{Mitigating harmful content and preserving platform integrity.}
AI-generated media increases the risk of deceptive and harmful content, such as deepfakes and misinformation that are difficult for users to identify~\cite[e.g.][]{associatedpress2024trump}. 
These materials can enable harassment, fraud, and manipulation, with real-world consequences for public safety and democratic processes~\cite[e.g.][]{wired2024aipoweredscams}.
At the same time, widespread generation risks saturating social feeds with synthetic content, crowding out authentic voices. 
Automated AIGC detection supports \textit{early intervention}, 
enabling platforms to flag or demote harmful media while preserving content diversity and platform integrity.

\paragraph{Detecting realistic AIGC.}
A challenge in AIGC detection lies in identifying highly realistic AI-generated content that closely mimics authentic content. 
As generative models improve, synthetic images and videos increasingly exhibit surface coherence that evade existing detectors~\cite{google_deepmind_veo_2024}. 
This underscores the need for evolving detection systems that go beyond shallow cues, instead leveraging semantic and contextual information for detection.

\section{Conclusion}
We propose a unified framework for detecting and explaining AI-generated content (AIGC) on social media. 
Our approach continuously curates in-the-wild multimodal data, leverages an efficient vision-language model for robust detection. 
Extensive experiments on public benchmarks and internal datasets show state-of-the-art detection performance, and large-scale deployment demonstrates tangible gains in user engagement, highlighting the practical impact of our system.

\section*{Limitations}
Despite strong empirical performance, our work has several limitations. 
First, while our continuous data collection pipeline mitigates distribution shift, it remains reactive to emerging generators and may lag behind new models or adversarial attacks specifically designed to evade detection. 
Second, although we train a small vision-language model for efficient inference, large-scale deployment still incurs nontrivial computational costs. 
Finally, the generated textual explanations aim to improve interpretability but do not guarantee faithful causal attribution, and may occasionally reflect model biases or overconfidence.

\bibliography{sample-base}

\appendix

\clearpage

\section{Prompt for AIGC Detection}

\begin{nicebox}
\small
\label{appendex:ifm_detection_prompt}
\textbf{Image-level detection Prompt:} \\
<image>\\ Does the image look real/fake?
\end{nicebox}

\begin{nicebox}
\small
\label{appendex:ifm_post_detection_prompt}
\textbf{Post-level detection Prompt:} \\
You are an Assistant to answer questions from users.
Your task is to evaluate the likelihood of a post being AI-generated. Specially, pay attention to the image included in the post.\\

<image>\\ 

BODY TEXT: \{body text\}\\
OCR TEXT: \{ocr text\}\\
TITLE TEXT: \{title text\}\\
VIDEO TRANSCRIPT: \{truncated video transcript\}\\

Given the post information, determine if the post is AI-generated or not. Answer in "yes" or "no".
\end{nicebox}

\newpage

\begin{nicebox}
\small
\label{appendex:reason_prompt}
\textbf{Detection + explanation prompt:}\\
You are an assistant designed to analyze images for potential AI-generated content by examining semantic clues, stylistic clues, and local visual artifacts.\\

<image>\\
\#\# Analysis Steps\\

1. **Semantic Clues Analysis**\\
\\
   - **Objective**: Identify dramatic or unlikely content that suggests the use of an AI generator.\\
   - **Output**: Provide a caption and reasoning for the scene, highlighting any elements that appear improbable or exaggerated.\\
\\
2. **Stylistic Clues Analysis**\\
\\
   - **Objective**: Detect waxy styles or overly high-quality features typical of generated images. Consider color, texture, and lighting.\\
   - **Output**: Offer a description of the image's style, noting any characteristics that deviate from natural or expected artistic styles.\\
\\
3. **Local Visual Artifacts Analysis**\\
\\
   - **Objective**: Identify imperfections and artifacts that are uncommon in real photos or traditional art. Consider physical artifacts (e.g., optical display issues, violations of physical laws, and spatial/perspective errors), structural artifacts (e.g., deformed objects, asymmetry, or distorted text), and distortion artifacts (e.g., color/texture distortion, noise/blur, artistic style errors, and material misrepresentation).\\
   - **Output**: Detail any local visual artifacts, explaining their nature and why they suggest synthetic generation.\\
\\
4. **Conclusion**\\
\\
    - After conducting the three-step analysis, you will provide a final assessment of the image's authenticity, considering the findings from each step. \\
    \\
Answer in the following format:\\
\\
<semantic>The image shows... </semantic>\\
<stylistic>The image is... </stylistic>\\
<local>The image has... </local>\\
<conclusion>AI-generated: Yes.</conclusion>
\end{nicebox}

\clearpage

\section{Ablation study}
\label{appendix:ablation}

\subsection{Ablations on Vision Encoders}
To understand the effectiveness of our vision encoder, we conducted ablation analysis comparing different variants:

\begin{itemize}[noitemsep, topsep=0pt]
    \item \textbf{CLIP vision encoder}: We replaced the vision encoder with a CLIP vision encoder trained on internal data. 
    \item \textbf{Larger input size}: We changed the input size from $448\times448$ to $896\times896$.
    \item \textbf{More visual tokens}: We changed the number of visual tokens from 256 to 1024.
\end{itemize}

We followed the same training and evaluation setup as before and trained each variant for only 1 epoch.

\begin{table}[t]
    \centering
    \footnotesize
    \begin{tabular}{p{3cm}cccc}
        \toprule
        \textbf{Method} & \multicolumn{2}{c}{\textbf{FakeClue}} & \multicolumn{2}{c}{\textbf{LOKI}} \\
        \cmidrule(lr){2-3} \cmidrule(lr){4-5}
        & \textbf{Acc $\uparrow$} & \textbf{F1 $\uparrow$} & \textbf{Acc $\uparrow$} & \textbf{F1 $\uparrow$} \\
        \midrule
        \ourmodel{}  & 0.929 & {0.945} & 0.793 & {0.833}   \\
        + {CLIP vision encoder} & 0.902 & 0.925 & 0.803 & 0.850 \\
        + {Larger input size} & 0.967 & 0.974 & 0.833 & 0.872 \\
        + {More visual tokens} & \textbf{0.973} & \textbf{0.979} & \textbf{0.859} & \textbf{0.884} \\
        \bottomrule
    \end{tabular}
    \caption{Impact of Perception Encoder and vision budget allocation on \ourmodel's performance \textit{when training for 1 epoch}. The Perception Encoder improves in-distribution accuracy compared to the smaller CLIP vision encoder but generalizes slightly worse on the LOKI dataset. Increasing vision input size or visual tokens significantly boosts performance.}
    \label{tab:ve-ablation}
\end{table}

\paragraph{Results.}
We first found that Perception Encoder does improve \ourmodel's performance under in-distribution setting, compared to the smaller CLIP vision encoder, though it generalizes slightly worse to the LOKI dataset (Table~\ref{tab:ve-ablation}).

Next, we found that allocating more budgets for the vision components -- either supporting larger input size or using more visual tokens -- significantly improves model performance when training for 1 epoch, indicating a trade-off between higher compute cost and fast training convergence.

\subsection{Ablations on Supervision Loss}
To evaluate the impact of different loss functions on model performance beyond standard SFT Language Modeling loss, we conducted ablations using the following loss variants:

\begin{itemize}[noitemsep, topsep=0pt]
    \item \textbf{BCE loss}: The standard binary cross-entropy loss for binary classification. We swap the language modeling head with a classification head. 
    \item \textbf{Focal loss}: A modified version of BCE loss that down-weights easy examples and focuses training on hard examples. 
    This is intended to improve model robustness and performance on imbalanced data.
\end{itemize}

We followed the same setups as before and trained each variant for two epochs.

\begin{table}[t]
    \centering
    \footnotesize
    \begin{tabular}{p{3cm}cccc}
        \toprule
        \textbf{Method} & \multicolumn{2}{c}{\textbf{FakeClue}} & \multicolumn{2}{c}{\textbf{LOKI}} \\
        \cmidrule(lr){2-3} \cmidrule(lr){4-5}
        & \textbf{Acc $\uparrow$} & \textbf{F1 $\uparrow$} & \textbf{Acc $\uparrow$} & \textbf{F1 $\uparrow$} \\
        \midrule
        \ourmodel, LM loss & 0.986 & 0.989 & 0.839 & 0.870 \\
        \ourmodel, BCE loss & 0.991 & 0.993 & 0.829 & 0.863 \\
        \ourmodel, focal loss & \textbf{0.991} &\textbf{0.993} & \textbf{0.845} & \textbf{0.875} \\
        \bottomrule
    \end{tabular}
    \caption{Performance of \ourmodel{} variants on FakeClue and LOKI datasets. BCE and focal loss variants show slightly improved performance compared to the standard SFT loss.}
    \label{tab:loss-ablation}
\end{table}

\paragraph{Results.}
We found that switching to the classification head can slightly improve \ourmodel's AIGC detection performance (Table~\ref{tab:loss-ablation}), with focal loss further improving model's performance on out-of-distribution settings where label imbalances differ.
This indicates that, for pure detection task without explanation or reasoning, using a classification head with focal loss will achieve the best result.

\subsection{Ablations on Visual Backbones and Data}
\label{appendix:ablations-visual}

To better understand the contributions of different design choices, we evaluate several ablation setups:  
\begin{itemize}[noitemsep, topsep=0pt]
    \item A variant without random text drop, testing whether having full access to text inputs affects performance.  
    \item A variant where the Perception Encoder is replaced by a CLIP-based vision encoder, testing sensitivity to the choice of vision backbone.  
    \item Variants that process 4-frame and 8-frame video inputs instead of the default setting (1-frame), testing extensions to multi-frame data.  
\end{itemize}

\clearpage

\begin{table}[t]
    \centering
    \footnotesize
    \begin{tabular}{p{4cm}cccc}
        \toprule
        \textbf{Method} & \textbf{Precision} & \textbf{Recall} \\
        \midrule
        \ourmodel  & 0.886 & 0.853 \\
        Embedding-based model   & 0.663     & 0.773  \\
        Adapter-based model   & 0.883     & 0.801  \\
        \midrule
        \multicolumn{3}{l}{\textbf{Ablations of \ourmodel}} \\
        No text drop  & 0.956 & 0.500 \\
        CLIP vision encoder  & 0.876 & 0.854 \\
        4-frame video inputs & 0.898  & 0.861  \\
        8-frame video inputs  & 0.904   &  0.849 \\
        \bottomrule
    \end{tabular}
    \caption{Comparison of precision and recall for different variants of \ourmodel{} in detecting AI-generated content. 
    Removing the random text drop drastically reduces recall to 0.500, indicating that models tend to infer based only on text content without text drop. 
    Replacing the Perception Encoder with the CLIP vision encoder results in slightly lower precision (0.876) but comparable recall (0.854).}
    \label{tab:aigc-results-compare}
\end{table}

\begin{table*}[t]
    \centering
    \footnotesize
    \begin{tabular}{lcccc}
        \toprule
        \textbf{Method} & \textbf{Valid Rate} & \textbf{Precision} & \textbf{Recall} & \textbf{Accuracy} \\
        \midrule
        SFT on binary classification & - & 0.889 & 0.785 & 0.949 \\
        SFT on 20k llama4-annotated data & 0.995 & 0.639 & 0.756 & 0.890 \\
        \midrule
        \multicolumn{5}{l}{\textit{Preference Alignment}} \\
        \midrule
        llama4-annotated data + 61K preference data (DPO) & 0.568 & - & - & - \\
        llama4-annotated data + 61K preference data (IPO) & 0.895 & 0.539 & 0.732 & 0.861 \\
        \midrule
        \multicolumn{5}{l}{\textit{SFT with rejection sampling}} \\
        \midrule
        llama4-annotated data + 61K rejection sampling data & 0.997 & 0.750 & 0.698 & 0.912 \\
        191K rejection sampling data & 0.997 & 0.745 & 0.687 & 0.910 \\
        191K rejection sampling data, llama3.2-11b & 0.995 & 0.695 & 0.558 & 0.888 \\
        \bottomrule
    \end{tabular}
    \caption{Evaluation results of various training and alignment methods on 200k examples for AI-generated content detection. Reasoning-enhanced fine-tuning alone yields decent performance and can be further improved by supervised fine-tuning (SFT) with rejection sampling data. 
    Despite improvements, gaps remain between reasoning and non-reasoning methods, indicating a trade-off between detection accuracy and generating human-interpretable reasoning traces.}
    \label{tab:results-200k}
\end{table*}

\begin{figure*}[t]
    \centering
    \includegraphics[trim=280pt 350pt 270pt 200pt, clip, width=\linewidth]{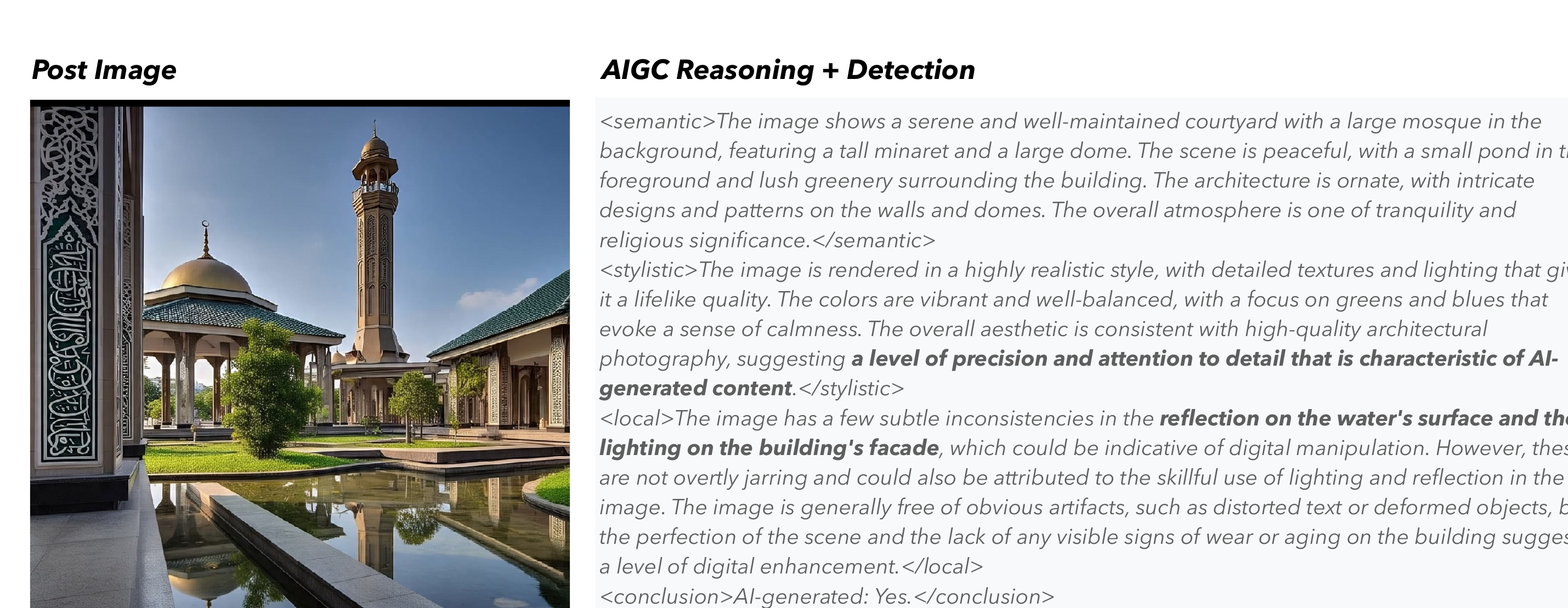}
    \caption{Example of AI-generated reasoning trace for AIGC detection. 
    The model provides semantic, stylistic, and local analyses. 
    This structured approach demonstrates the model’s ability to generate plausible, human-interpretable reasoning when identifying AI-generated content.}\label{fig:reasoning_example}
\end{figure*}

\section{Reasoning about AI-generated Content on Social Media}
\label{sec:experiment-internal-explain}

We then explore \ourmodel's ability to reason about AI-generated content on social media.
This is to understand how well \ourmodel{} can generate plausible reasoning traces before making a final judgment.

\paragraph{Setups.}
We first curated 20k reasoning traces from \texttt{Llama4-Maverick} with a prompt (Appendix~\ref{appendex:reason_prompt}) that requires models to provide analysis on image semantics, style, and local visual artifacts (Figure~\ref{fig:reasoning_example}). 
We then fine-tuned \ourmodel{} with the 20k reasoning-enhanced data.

To increase input coverage, we performed rejection sampling with this fine-tuned checkpoint on our curated \textbf{AIGC Fine-tuning} dataset, with 8 samples per input.
We curated 61k pairwise preference data and 61K SFT data on hard examples this way, by verifying each completion's final conclusion against the label ground truth.
We further curated 191k SFT data by collecting all examples with at least one correct response.
We explored various post-training methods (rejection sampling finetuning, direct preference optimization, group relative policy optimization) on this dataset.

\paragraph{Results.}

Overall, we found \ourmodel{} achieves strong performance in detecting AI-generated content while generating plausible reasoning traces (Table~\ref{tab:results-200k}).
Breaking down the results, we observed the following:

First, we found that \textbf{reasoning-enhanced fine-tuning alone yields decent performance}, with a precision of 0.639 and a recall of 0.756. 
This can be further improved by SFT with rejection sampling data, which expose the model to more diverse samples. 
However, gaps remain between reasoning and non-reasoning methods, indicating a trade-off for generating human-interpretable traces.

Second, we found that \textbf{preference alignment methods such as DPO and IPO show mixed results} -- models often exhibit serious reward hacking issues with, leading to worse format following and increased repetition that produce malformed outputs.

Third, we found that larger models like \texttt{Llama3.2-11b} do not yield better results, similar to what we have observed in Section~\ref{sec:experiment-benchmark}, likely due to their limited exposure to social media data.

\subsection{Reasoning faithfulness}
We conducted an exploratory analysis to estimate the reasoning error rate.
We mark the model-provided reasoning as correct if \emph{at least one} of the multiple reasons generated for a sample is valid, since the model typically outputs several supporting reasons per example.

\paragraph{Results.}
For images, the reasoning error rate is below 5\%; the dominant failure mode is not missing genuine artifacts, but hallucinating additional ones (e.g., attributing ``unnatural textures'' to artifact-free regions).
For videos, the reasoning error rate is higher (above 10\%), largely due to frame sampling limitations that amplify hallucinations on fast-paced content.

\paragraph{Examples.}
We share a few anonymized examples (images omitted due to privacy concerns) to demonstrate when we observe the reasoning to be (not) faithful:

\begin{itemize}[noitemsep, topsep=0pt, leftmargin=15pt]
    \item \textbf{Detecting unrealistic synthetic scene}:\textit{ ``The entire scene appears generated from scratch. The dog is anthropomorphized (standing at a bank counter, talking), and the human character has the characteristic smooth, slightly plastic look of AI-generated humans.''} (Faithful)
    \item \textbf{Detecting garbled text}: \textit{``The images exhibit definitive markers of AI generation. Most notably, whenever text is supposed to be visible within the images themselves (such as the pages of the Bible or the scroll), it is completely garbled, nonsensical pseudo-script. There are also minor anatomical inconsistencies in the hands across various images.''} (Faithful)
    \item \textbf{Hallucinating about ``unnatural motions''}: \textit{However, there are subtle signs of AI-generated motion: the woman’s hand movements show slight morphing and instability — her fingers appear to blend or stretch unnaturally as she gestures.} (Unfaithful)
\end{itemize}

\clearpage

\section{Error analysis}
\label{appendix:error-analysis}
We summarize common error patterns observed during development and evaluation, along with their likely causes and mitigation.

\paragraph{Over-reliance on textual cues.}
A frequent false positive occurs when the text contains surface-level ``AI'' indicators (e.g., explicit mentions of AI tools, prompts, or model names) while the content itself is human-written.
This suggests the model can shortcut by keying on lexical cues rather than grounded artifacts.
To reduce this behavior, we apply aggressive text dropout (90\%) during training, forcing the model to rely more on visual and multimodal signals.

\paragraph{Out-of-distribution categories.}
The model degrades on categories that differ substantially from the training distribution, such as screenshots of PDFs.
Errors here are primarily false negatives: subtle AIGC artifacts are missed when the input departs from common social-media-style images.
A practical takeaway is that deploying the model beyond the intended domain should be paired with domain-specific data curation (e.g., document-centric AIGC examples) and/or targeted adaptation.

\paragraph{Large images with partially AI-generated regions.}
Another recurring failure mode involves high-resolution images that are mostly real but contain a small AI-generated region (e.g., a document image with an inserted AIGC illustration).
When the manipulated region is spatially small, a global representation can underweight it, leading to false negatives.
We partially mitigate this by allocating more compute budget to the vision components (e.g., higher-resolution features / more visual tokens), which improves sensitivity to localized artifacts.

\paragraph{Videos with partially AI-generated frames.}
We also observe failures on video inputs where only a subset of frames are AI-generated.
When evidence is temporally sparse, frame sampling and temporal aggregation can wash out the signal, yielding false negatives.
This suggests that robust video detection likely requires adaptive sampling strategies that increase coverage around suspicious segments.

\end{document}